\documentclass[runningheads]{llncs}
\usepackage{graphicx}
\usepackage{amsmath,amssymb} 
\usepackage{color}

\begin{document}
\title{Knowing Where to Look? Analysis on Attention of Visual Question Answering System} 

\titlerunning{Analysis on Attention of Visual Question Answering System}
%
\author{Wei Li\inst{1,2\thanks{This work was done while the author was a research intern in ByteDance AI Lab.}}\and
Zehuan Yuan\inst{2} \and
Xiangzhong Fang\inst{1} \and
Changhu Wang\inst{2}}
%
\authorrunning{Wei li et al.}
%

\institute{Shanghai Jiao Tong University, Shanghai \\
\email{\{liweihfyz,xzfang\}@sjtu.edu.cn} \and
ByteDance AI Lab \\
\email{\{levi.li,yuanzehuan,wangchanghu\}@bytedance.com}}
\maketitle

\begin{abstract}

Attention mechanisms have been widely used in Visual Question Answering (VQA) solutions due to their capacity to model deep cross-domain interactions. Analyzing attention maps offers us a perspective to find out limitations of current VQA systems and an opportunity to further improve them. In this paper, we select two state-of-the-art VQA approaches with attention mechanisms to study their robustness and disadvantages by visualizing and analyzing their estimated attention maps. We find that both methods are sensitive to features, and simultaneously, they perform badly for counting and multi-object related questions. We believe that the findings and analytical method will help researchers identify crucial challenges on the way to improve their own VQA systems.

\end{abstract}

\section{Introduction}
Visual question answering (VQA) attracts increasing attentions in both computer vision and natural language processing community. The goal of VQA is to answer questions based on the information of any given image. 
As deep learning witnessed a series of remarkable success in artificial intelligence, VQA also made tremendous progress \cite{Fukui2016MultimodalCB,yu2018beyond,Kim2017} over past few years such as several benchmark datasets, e.g., 
VQA 2.0 \cite{balanced_vqa_v2}, CLEVR \cite{2016arXiv161206890J} and Visual Genome \cite{krishnavisualgenome}, and tons of approaches, e.g., MFB~\cite{yu2018beyond} and BAN~\cite{2018kim}.

VQA is usually formulated as a classification task with different answers as candidate categories. The current mainstream pipeline is to firstly extract image and question representations with Convolutional Neural Network and Recurrent Neural Network, respectively. Then, a lot of fusion methods such as early fusion \cite{2015zhou} and bilinear pooling \cite{yu2018beyond,Kim2017,Fukui2016MultimodalCB,2018kim} are adopted to combine two-stream features. In addition, attention is playing an increasingly important role as the mechanism encourages deep cross-domain interactions without introducing substantial parameters. 
There are two main branches to add attention to VQA system: {\it uni-attention} and {\it co-attention}. {\it Uni-attention} merely considers question-guided visual attentions. In contrast, {\it co-attention} additionally takes image-guided question attentions into account to jointly model the multimodal correlations \cite{2016co_attention,Nguyen_2018_CVPR,2018kim}.


Although much progress has been made, few works lie on deep analysis on the influence of different attention mechanisms. In this paper, we dive into two state-of-the-art methods: multi-model factorized pooling (MFB) \cite{yu2018beyond} and bilinear attention network (BAN) \cite{2018kim} to discover their inherent limitations. Both methods adopt the popular bilinear pooling to perform multimodal fusion. However, MFB only performs question-guided visual attention ({\it uni-attention}) while BAN extends {\it co-attention} into bilinear attention to enable more image and language interactions.
We conduct all our experiments on VQA 2.0 dataset with a more balanced answer distribution than VQA 1.0 \cite{balanced_binary_vqa} and Visual Genome dataset. 
In addition, it covers more relations of real-world objects compared with CLEVR dataset full of synthetic images.
In order to make a deeper understanding of both methods, we propose to directly delve into their attention maps. Observing whether estimated attentions relate to real answers could reflect the robustness and limitations of corresponding approaches.

To summarize, we present three key observations after thorough experiments on both approaches:
\begin{itemize}
\item [-] 
The performance is sensitive to selected features. Representations based on object proposals are better than image-level features.
\item [-]
Attention distribution becomes much more inaccurate for questions related to multiple objects.
\item [-]
Counting problem is not well solved by soft attention mechanism.
\end{itemize}

In terms of each observation, we also analyze main reasons behind these phenomenons and claim that similar limitations probably exist in most of methods with attention mechanisms. We believe that these findings will inspire researchers to design more effective methods. Furthermore, our analytical method is hopeful to offer researchers an opportunity to identify potential roadblocks when debugging their VQA systems.

\section{Multimodal Factorized Bilinear Pooling Revisited}
Since bilinear pooling \cite{tenenbaum1997separating} allows abundant multimodal cross-channel interactions, the fusion method has been widely used in VQA systems compared to simple summation and concatenation operators. To further reduce the number of parameters in bilinear pooling, multimodal factorized bilinear Pooling (MFB) \cite{yu2018beyond} decomposes the weight matrix as two low-rank matrices. 

Specifically, given a question vector $x\in \mathbb{R}^m$ and an image feature vector $y\in\mathbb{R}^n$, each output channel of MFB pooling is formulated as:
\begin{equation}
\text{pool}(x, y)_i=x^T \textbf{W}_i y+b_i=x^T \textbf{U}_i \textbf{V}_i^T y+b_i=\mathbb{I}(\textbf{U}_i^T x \circ \textbf{V}_i^T y)+b_i
\end{equation}
where $\mathbb{I} \in\mathbb{R}^k$ is a vector of all elements ones, $\textbf{W}_i\in\mathbb{R}^{m\times n}$ is the weight matrix and $\textbf{U}_i\in\mathbb{R}^{m\times k}$ and $V_i\in\mathbb{R}^{n\times k}$ are two factorized matrices. 

The whole pipeline of MFB for VQA can be summarized as follows. First, an overall question representation $\hat{x}\in\mathbb{R}^m$ is obtained by a {\it self-attention} manner with weights $\alpha^x$. Then, the weighted question feature guilds the visual attention on the image as follows:
\begin{equation}
\alpha^y = softmax(\{\textbf{W}_p^T \text{pool}(\hat{x}, y_j)\}), \hat{y} = \sum_j \alpha_j^y y_j
\end{equation}
where $y_j$ is an image feature vector and $\textbf{W}_p\in\mathbb{R}^{m\times1}$. Finally, attention weighted language feature $\hat{x}$ and visual feature $\hat{y}$ are fused together as $f=\text{pool}(\hat{x},\hat{y})$ for further prediction.

\section{Bilinear Attention Revisited}
Co-attention based model jointly integrates question-guided visual attention and visual-guided question attention together. To further consider every pair of multimodal features, BAN \cite{2018kim} extends co-attention into bilinear attention. The fused feature can be defineds as:
\begin{equation}
    f_i=(\textbf{X}^T {\tilde{\textbf{U}}})_i^T A (\textbf{Y}^T \tilde{\textbf{V}})_i
\end{equation}
where $\tilde{\textbf{U}} \in\mathbb{R}^{m\times k}$, $\tilde{\textbf{V}}\in\mathbb{R}^{n\times k}$, $\textbf{X}\in\mathbb{R}^{m\times \theta}$, $\textbf{Y}\in\mathbb{R}^{n\times \gamma}$, and $A\in\mathbb{R}^{\theta\times \gamma}$ is the bilinear attention map that sums to 1 as follows:
\begin{equation}
    A=softmax((\mathbb{I}\cdotp p^T) \circ X^T \textbf{U})\textbf{V}^T Y)
\end{equation}
where $\mathbb{I}\in\mathbb{R}^\theta$ is a vector with all elements ones, $p \in\mathbb{R}^k$, and $softmax$ is applied element-wisely. Then the fused feature $f$ can be used for further classification.

MFB and BAN represent popular attempts in {\it uni-attention} and {\it co-attention} directions, respectively. 
A thorough analysis for both methods is also expected to shed light on similar limitations of other approaches with attention mechanisms. 

\section{Deep Study}

In this section, we will present detailed analysis for our key observation results. As shown above, we investigate MFB \cite{yu2018beyond} and BAN \cite{2018kim} to make a thorough study. All experiments are conducted on VQA2.0 benchmark, where we train on {\it train} split with 82,783 images and 443,757 questions, and evaluate on {\it val} split with 40,503 images and 214,354 questions totally. Each question is annotated with 10 answers by crowdsourcing. In order to give an intuitive demonstration, we report visualizations of image attention vectors $\alpha^y$ in MFB and the bilinear attention maps $A$ in BAN.

\begin{figure}
\centering
\includegraphics[height=5cm]{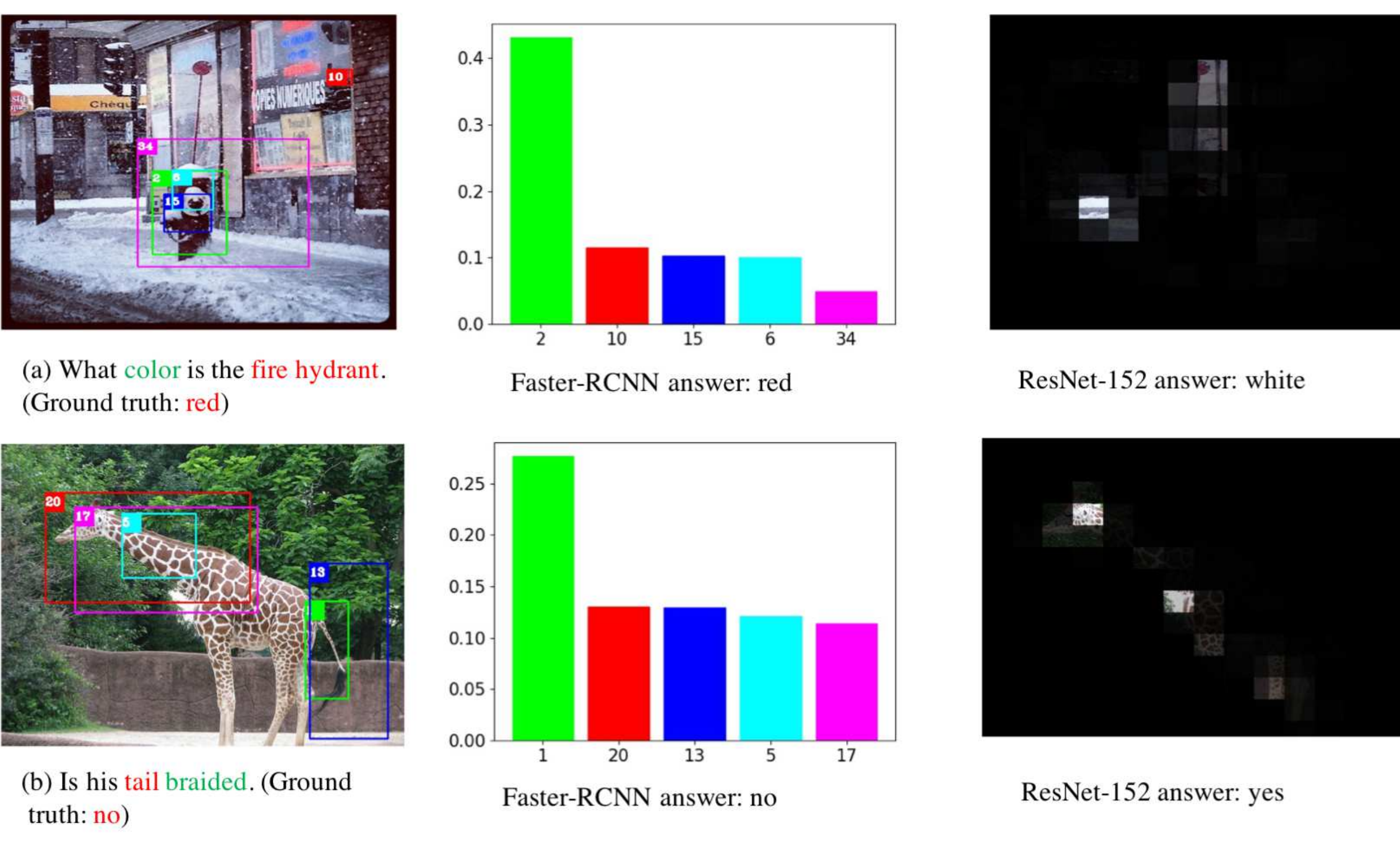}
\caption{Visualization of MFB with different visual features. From left to right are the original images, the MFB attention weights of Faster-RCNN proposals and the MFB attention map of the ResNet-152 feature map. The most salient boxes (numbered in the top-left corner of each bounding box and x-axis of the grids) are visualized in both images.}
\label{figure:2}
\end{figure}

\subsection{Object feature \& Image feature}

Visual object features have been proven effective in VQA task \cite{Teney_2018_CVPR,2018kim} compared with image-level features.  However, the reason behind the performance gain has not been well investigated. In this work, we delve deeper into this from the attention perspective.

In our experiments, we select top-36 Faster-RCNN proposals \cite{renNIPS15fasterrcnn} and ResNet-152 last feature map before {\it pool5} \cite{He_2016_CVPR} as object features ($36\times 2,048$) and image features ($196\times 2,048$), respectively. We set the batch size to 64 and the dimension of hidden states to 1024 in BAN. To simplify experiments, we do not integrate counting module \cite{zhang2018vqacount}. Unlike the original implementation, we augment 300-dimensional random initialized word embedding instead of 300-dimensional computed word embedding to each 300-dimensional Glove word embedding. The performance comparison on the VQA 2.0 validation set is shown in Table~\ref{table:1}. Unsurprisingly, we achieve better performance with object features for both methods compared with image-level features. In addition, we found that a more accurate attention distribution can be obtained for object features compared with image features. For example in Fig.~\ref{figure:2}, given a question about fire hydrant, we can see that MFB with object proposals focuses on the correct entity while image-level representation directs attentions to snow regions. Due to the inaccurate attention distribution, the model with image features predicts a wrong answer, \emph{white}.  Similarly when ``\textbf{Is his tail braided?}'' is asked, the tail proposal is highlighted for the method with object-level representations as opposed to arbitrary emphasis with a single feature map.

\setlength{\tabcolsep}{2pt}
\begin{table}
\begin{center}
\caption{Detailed performance comparison on VQA 2.0 validation set}
\label{table:1}
\begin{tabular}{llllll}
\hline\noalign{\smallskip}
Feature type & Methods & Overall & Other & Number & Yes/No\\
\noalign{\smallskip}
\hline
\noalign{\smallskip}
\hline
ResNet-152 feature map  & MFB\cite{yu2018beyond} & 60.94 & 52.93 & 38.48 & 79.28\\
~ & BAN\cite{2018kim} & 59.52 & 51.19 & 38.92 & 77.64\\
Faster-RCNN proposals & MFB\cite{yu2018beyond} & 65.19 & 57.17 & 44.37 & 82.98\\
~ & BAN\cite{2018kim} & 64.3 & 55.7 & 45.45 & 82.16\\
\hline
\end{tabular}
\end{center}
\end{table}
\setlength{\tabcolsep}{1.4pt}

Although it is difficult to measure the negative effect of features quantitatively on attention maps over the entire dataset, we hypothesize that inaccurate attention maps take a large amount of responsibility for decline in performance.

We analyse that object proposals have much more specific semantic meanings compared with feature maps and thus the corresponding relations between words and visual features are easier to learn, which leads to a more accurate attention distribution and further performance boost.

\subsection{Single object \& Multiple objects}

\begin{figure}
\centering
\includegraphics[height=6.75cm]{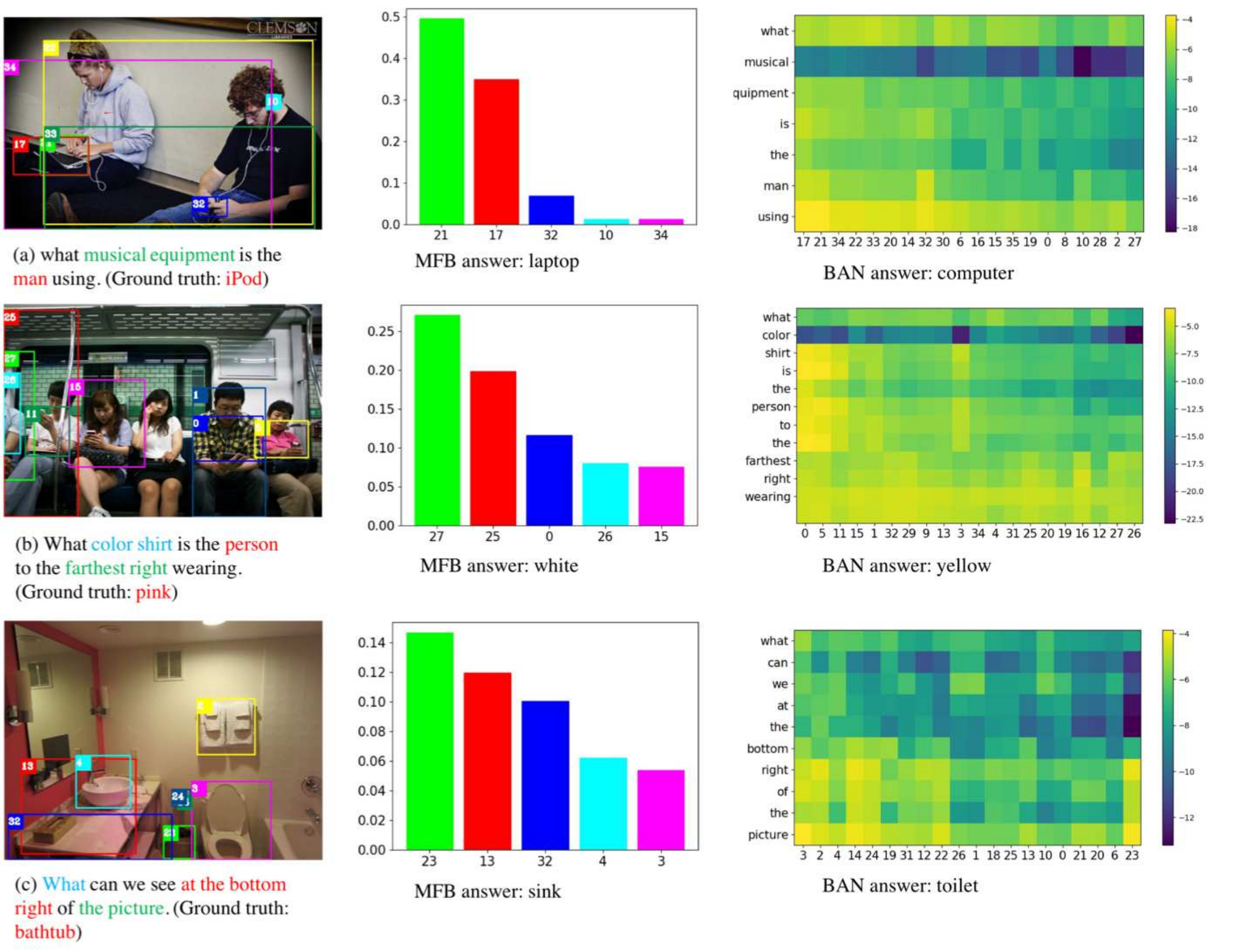}
\caption{Visualization of MFB and BAN on questions related to multiple objects. From left to right are the original images, MFB attention vectors and BAN bilinear attention maps. The most salient boxes (numbered in the top-left corner of each bounding box and x-axis of the grids) are visualized in both images.}
\label{figure:3}
\end{figure}

Based on how many objects are necessary to infer final answers, questions in VQA2.0 can be roughly divided into single object, e.g., ``what is the color of the dog?'' and multiple objects, e.g., ``what color is the book on the desk?''. In our experiment, we conduct the comparison for both kinds of questions. The observation shows that the attention distribution is much more inaccurate for questions related to multiple objects. For example in Fig.~\ref{figure:3}, both models incorrectly focus on the laptop used by the woman in (a), which implies that the relation between the woman and the laptop are not well captured and modeled. Additionally, relative positions are not well integrated by both models. We can see in Fig.~\ref{figure:3}, both models make predictions (\emph{white} and \emph{yellow}) based on the person on the left and the person on the middle respectively in (b). In a word, the estimated attention maps cannot learn relative positions. Moreover, spatial locations are crucial to infer the \emph{what} question in (c). Both models concentrate on the wrong objects in other positions, e.g, sink and toilet.

It is worth noting that current attention mechanisms learn attention distributions by only comparing visual and question representations and object features ignore their own locations in images.

However, without well-captured object relations or position information, models are unable to set these visually or semantically similar objects apart when the questions are related to multiple objects or multiple instances exist in an image. The confusion causes an inaccurate attention distribution which leads to a significant accuracy drop between single-object questions and those with multiple objects, which constitutes the main hurdle for current VQA systems.

In order to reduce the performance gap, it could be a crucial step to explicitly consider object relation and position. In particular, graph-based neural networks might be an effective way to handle unstructured object correlations \cite{xu2017scenegraph,Liu_2018_CVPR}. Object relations modeling is still an open question and worth further explorations.

\subsection{Counting problem}
Counting problem is a special case of questions related to multiple objects. As mentioned in \cite{zhang2018vqacount}, due to that soft-attention mechanism normalizes the attention weights, which leads to the loss of counting-related information. Soft attention is replaced by the gate strategy in \cite{zhang2018vqacount} and then overlapping object proposals are processed in a differentiable manner.

\begin{figure}
\centering
\includegraphics[height=6.75cm]{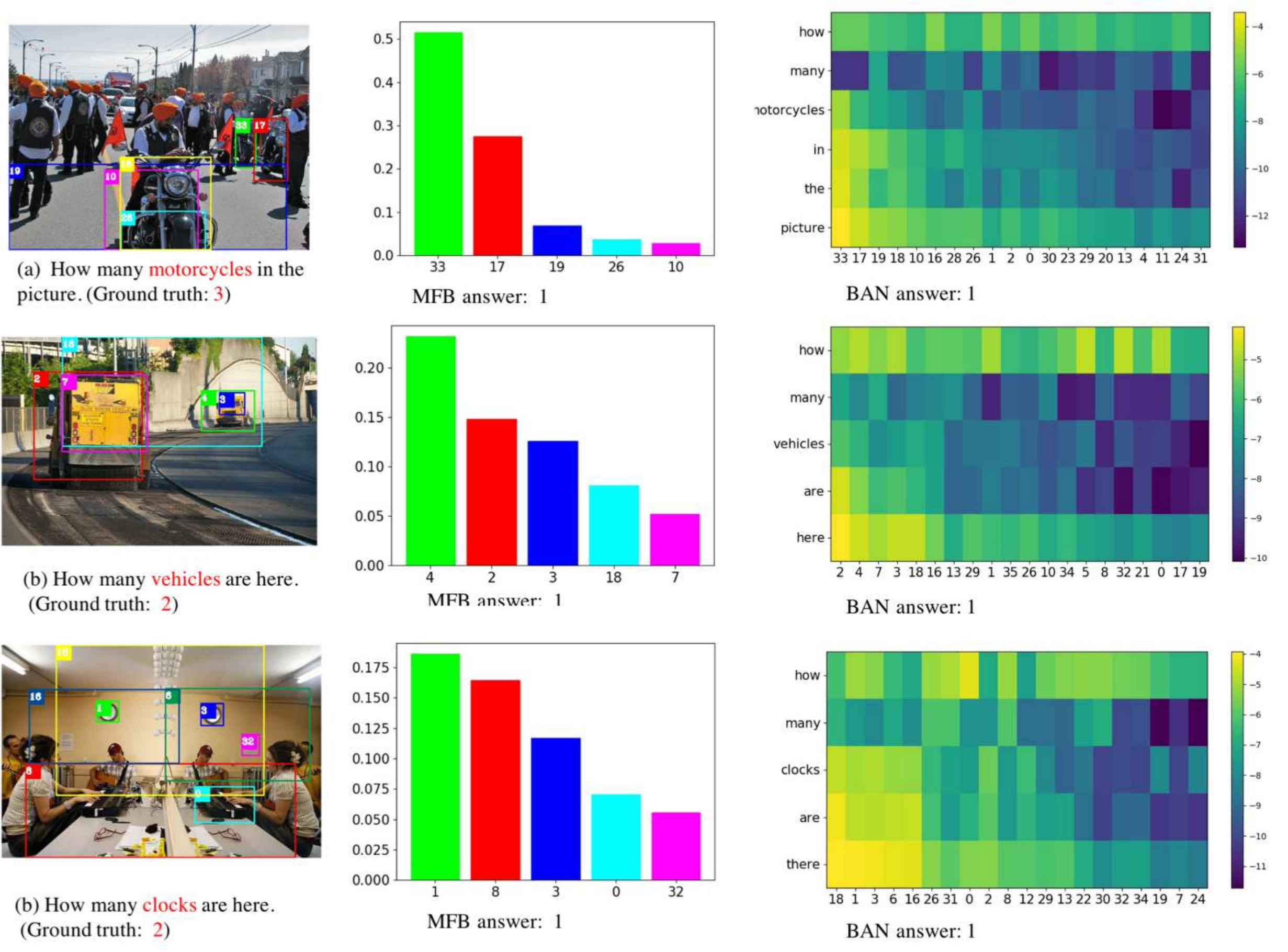}
\caption{Visualization of MFB and BAN on counting problems. From left to right are the original images, MFB attention vectors and BAN bilinear attention maps. The most salient boxes (numbered in the top-left corner of each bounding box and x-axis of the grids) are visualized in both images. Both models give the wrong answer, 1.}
\label{figure:4}
\end{figure}

In this work, we show that poor results can also be obtained even with an accurate attention distribution. For example in Fig.~\ref{figure:4}, both models focus their attention on multiple detected objects, namely, motorcycles in (a), vehicles in (b) and clocks in (c). However, detected objects are obviously visually similar and thus the weighted average of these visual features is probably similar to one of them, which means cues for counting are lost during soft attention process regardless of attention distributions. The limitations probably exist in a large amount of VQA systems. Therefore, in order to improve the counting performance essentially, additional structures or more flexible attention mechanisms might be needed. 

\section{Conclusions}
To facilitate further research on the VQA task, we delve into two state-of-the-art methods MFB \cite{yu2018beyond} and BAN \cite{2018kim} on VQA 2.0 dataset by visualizing and analysing their estimated attention maps. We form three main observations. Firstly, the performance improvement with Faster-RCNN proposals is probably related to a more accurate attention distribution. Second, the attention distribution is much more inaccurate for questions related to multiple objects. Finally, counting problem is not well solved by soft attention mechanism due to the attention weight normalization. We believe that these observation results can help future VQA research and analysing attention maps will also assist researchers to debug their own VQA systems.
\bibliographystyle{splncs04}
\bibliography{mybibliography}

\begin{thebibliography}{10}
\providecommand{\url}[1]{\texttt{#1}}
\providecommand{\urlprefix}{URL }
\providecommand{\doi}[1]{https://doi.org/#1}

\bibitem{Fukui2016MultimodalCB}
Fukui, A., Park, D.H., Yang, D., Rohrbach, A., Darrell, T., Rohrbach, M.:
  Multimodal compact bilinear pooling for visual question answering and visual
  grounding. In: Conference on Empirical Methods in Natural Language Processing
  (EMNLP) (2016)

\bibitem{balanced_vqa_v2}
Goyal, Y., Khot, T., Summers{-}Stay, D., Batra, D., Parikh, D.: Making the {V}
  in {VQA} matter: Elevating the role of image understanding in {V}isual
  {Q}uestion {A}nswering. In: Conference on Computer Vision and Pattern
  Recognition (CVPR) (2017)

\bibitem{He_2016_CVPR}
He, K., Zhang, X., Ren, S., Sun, J.: Deep residual learning for image
  recognition. In: Conference on Computer Vision and Pattern Recognition (CVPR)
  (2016)

\bibitem{2016arXiv161206890J}
{Johnson}, J., {Hariharan}, B., {van der Maaten}, L., {Fei-Fei}, L., {Zitnick},
  C.L., {Girshick}, R.: {CLEVR: A Diagnostic Dataset for Compositional Language
  and Elementary Visual Reasoning}. arXiv preprint arXiv:1612.06890  (2016)

\bibitem{2018kim}
{Kim}, J.H., {Jun}, J., {Zhang}, B.T.: {Bilinear Attention Networks}. arXiv
  preprint arXiv:1805.07932  (2018)

\bibitem{Kim2017}
Kim, J.H., On, K.W., Lim, W., Kim, J., Ha, J.W., Zhang, B.T.: {Hadamard Product
  for Low-rank Bilinear Pooling}. In: International Conference on Learning
  Representations (ICLR) (2017)

\bibitem{krishnavisualgenome}
Krishna, R., Zhu, Y., Groth, O., Johnson, J., Hata, K., Kravitz, J., Chen, S.,
  Kalantidis, Y., Li, L.J., Shamma, D.A., Bernstein, M.S., Fei-Fei, L.: Visual
  genome: Connecting language and vision using crowdsourced dense image
  annotations. International Journal of Computer Vision (IJCV)
  \textbf{123}(1),  32--73 (2017)

\bibitem{Liu_2018_CVPR}
Liu, Y., Wang, R., Shan, S., Chen, X.: Structure inference net: Object
  detection using scene-level context and instance-level relationships. In:
  Conference on Computer Vision and Pattern Recognition (CVPR) (2018)

\bibitem{2016co_attention}
{Lu}, J., {Yang}, J., {Batra}, D., {Parikh}, D.: {Hierarchical Question-Image
  Co-Attention for Visual Question Answering}. arXiv preprint arXiv:1606.00061
  (2016)

\bibitem{Nguyen_2018_CVPR}
Nguyen, D.K., Okatani, T.: Improved fusion of visual and language
  representations by dense symmetric co-attention for visual question
  answering. In: Conference on Computer Vision and Pattern Recognition (CVPR)
  (2018)

\bibitem{renNIPS15fasterrcnn}
Ren, S., He, K., Girshick, R., Sun, J.: Faster {R-CNN}: Towards real-time
  object detection with region proposal networks. In: Advances in Neural
  Information Processing Systems (NIPS) (2015)

\bibitem{tenenbaum1997separating}
Tenenbaum, J.B., Freeman, W.T.: Separating style and content. In: Advances in
  Neural Information Processing Systems (NIPS) (1997)

\bibitem{Teney_2018_CVPR}
Teney, D., Anderson, P., He, X., van~den Hengel, A.: Tips and tricks for visual
  question answering: Learnings from the 2017 challenge. In: Conference on
  Computer Vision and Pattern Recognition (CVPR) (2018)

\bibitem{xu2017scenegraph}
Xu, D., Zhu, Y., Choy, C., Fei-Fei, L.: Scene graph generation by iterative
  message passing. In: Conference on Computer Vision and Pattern Recognition
  (CVPR) (2017)

\bibitem{yu2018beyond}
Yu, Z., Yu, J., Xiang, C., Fan, J., Tao, D.: Beyond bilinear: Generalized
  multimodal factorized high-order pooling for visual question answering. IEEE
  Transactions on Neural Networks and Learning Systems (99),  1--13 (2018)

\bibitem{balanced_binary_vqa}
Zhang, P., Goyal, Y., Summers{-}Stay, D., Batra, D., Parikh, D.: {Y}in and
  {Y}ang: Balancing and answering binary visual questions. In: Conference on
  Computer Vision and Pattern Recognition (CVPR) (2016)

\bibitem{zhang2018vqacount}
Zhang, Y., Hare, J., Pr\"ugel-Bennett, A.: Learning to count objects in natural
  images for visual question answering. In: International Conference on
  Learning Representations (ICLR) (2018)

\bibitem{2015zhou}
{Zhou}, B., {Tian}, Y., {Sukhbaatar}, S., {Szlam}, A., {Fergus}, R.: {Simple
  Baseline for Visual Question Answering}. arXiv preprint arXiv:1512.02167
  (2015)

\end{thebibliography}
\end{document}